\pdfoutput=1

\documentclass[11pt]{article}

\usepackage[]{EMNLP2022}

\usepackage{times}
\usepackage{latexsym}

\usepackage[T1]{fontenc}

\usepackage[utf8]{inputenc}

\usepackage{microtype}

\usepackage{enumitem}
\usepackage{bookmark}
\usepackage{bm}
\usepackage{amsfonts}
\usepackage{subfigure}
\usepackage{graphicx}
\usepackage{booktabs}
\usepackage{makecell}
\usepackage{xcolor}
\usepackage{amsmath}
\usepackage{arydshln}
\usepackage{multirow}
\usepackage{stmaryrd}
\usepackage{color}
\usepackage{makecell}

\title{Mask-then-Fill: A Flexible and Effective Data Augmentation Framework for Event Extraction}

\author{ Jun Gao\textsuperscript{1,3}\thanks{\;\;Work done when Jun Gao was interning at 4Paradigm} \enskip Changlong Yu\textsuperscript{4}\enskip Wei Wang\textsuperscript{5}\enskip Huan Zhao\textsuperscript{6 }\enskip Ruifeng Xu\textsuperscript{1,2,3}\thanks{\;\;Corresponding author}\\
\normalsize \textsuperscript{1}Harbin Institute of Technology (Shenzhen)\quad \textsuperscript{2}Peng Cheng Laboratory\quad  \\
\normalsize \textsuperscript{3}Guangdong Provincial Key Laboratory of Novel Security Intelligence Technologies\\
\normalsize \texttt{imgaojun@gmail.com}\quad \texttt{xuruifeng@hit.edu.cn}\\
\normalsize   \textsuperscript{4}HKUST, Hong Kong, China\quad \textsuperscript{5}Tsinghua University \quad \textsuperscript{6}4Paradigm. Inc. \\
\normalsize \texttt{cyuaq@cse.ust.hk} \quad \texttt{weiwangorg@163.com} \quad \texttt{zhaohuan@4paradigm.com}\\
}

\begin{document}
\maketitle
\begin{abstract}
  We present \textbf{Mask-then-Fill}, a flexible and effective data augmentation framework for event extraction. Our approach allows for more flexible manipulation of text and thus can generate more diverse data while keeping the original event structure unchanged as much as possible. Specifically, it first randomly masks out an adjunct sentence fragment and then infills a variable-length text span with a fine-tuned infilling model. The main advantage lies in that it can replace a fragment of arbitrary length in the text with another fragment of variable length, compared to the existing methods which can only replace a single word or a fixed-length fragment. On trigger and argument extraction tasks, the proposed framework is more effective than baseline methods and it demonstrates particularly strong results in the low-resource setting. 
  Our further analysis shows that it achieves a good balance between diversity and distributional similarity.
\end{abstract}

\section{Introduction}
Event Extraction~(EE), which aims to extract triggers with specific types and their arguments from unstructured texts, is an important yet challenging task in natural language processing.
In recent years, deep learning methods have emerged as one of the most prominent approaches for this task~\citep{Nguyen2019OneFA,Lin2020AJN,du2020event,Paolini2021StructuredPA,lu2021text2event,lou2022translation}. However, they are notorious for requiring large labelled data, which limits the scalability of EE models. Annotating data for EE is usually costly and time-consuming, as it requires expert knowledge. 
One possible solution is to leverage data augmentation~(DA)~\citep{simard1998transformation}.


Existing DA methods for NLP can be broadly classified into two types: (1) the first is to augment training data by modifying existing examples~\citep{Sennrich2016ImprovingNM,csahin2019data,dai2020analysis,Wei2019EDAED}, and (2) the second is to generate new data by estimating a generative process and sample from it~\citep{anaby2020not,quteineh2020textual,Yang2020GenerativeDA,Ye2022ZeroGenEZ}. 
Since the EE task requires DA methods to generate augmented samples and token-level labels jointly, the second type of DA method is inapplicable here. In this study, we mainly focus on the first type of method. 



\begin{figure} 
  \centering
  \includegraphics[width=0.9\linewidth]{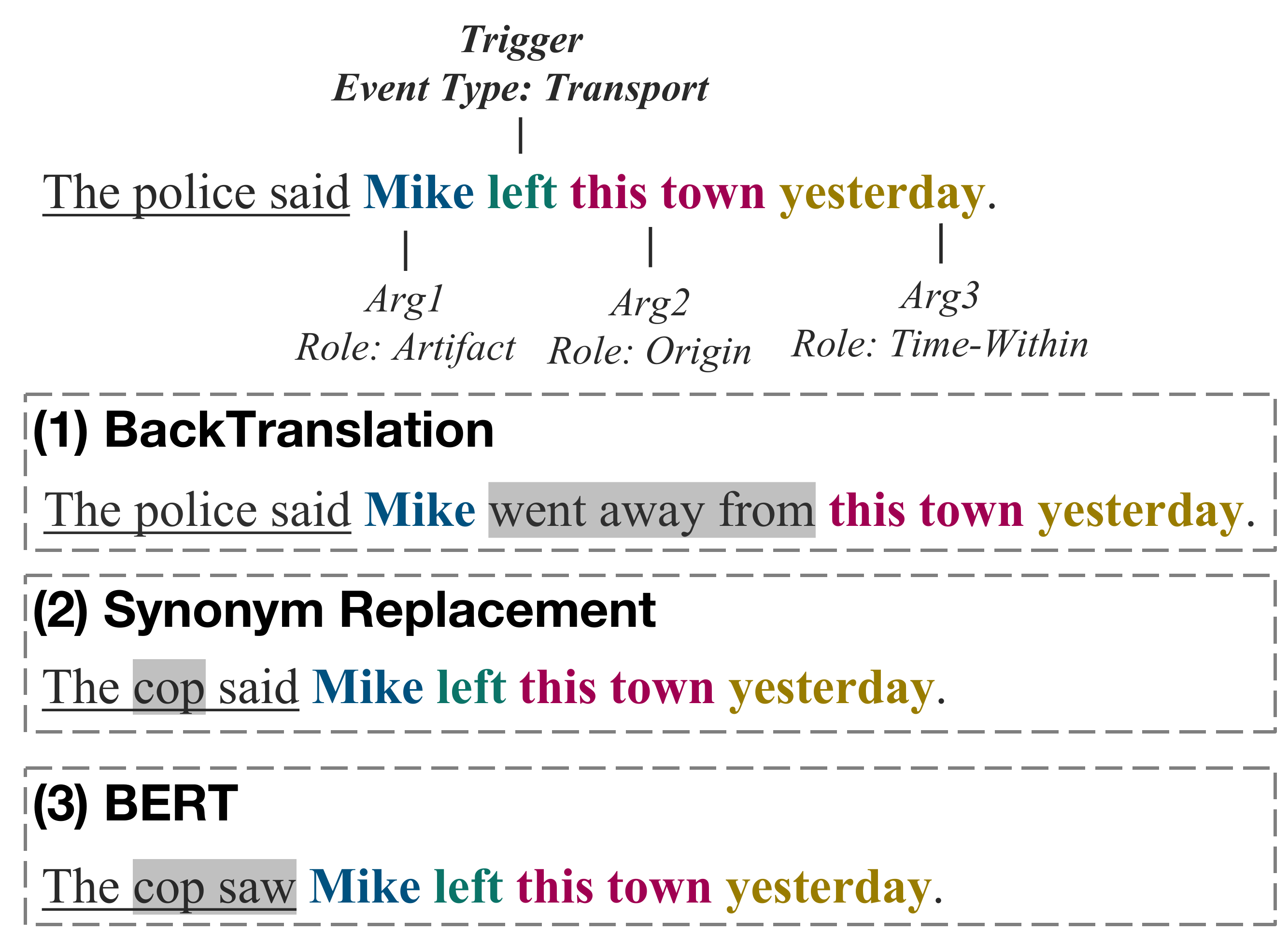}
  \caption{Visualization of three different data augmentation methods applied to a sentence containing a ``Transport'' event. Spans marked with different colors are event triggers and arguments. The parts of the augmented sample that differ from the original are colored in gray.
  \textbf{Backtranslation}~\citep{xie2020unsupervised} translates the input sentence into another language and back to the original. \textbf{Synonym Replacement}~\citep{dai2020analysis} and \textbf{BERT}~\citep{yang2019exploring} replace words in the sentence.}
  \vspace{-1.5em}
  \label{fig:example}
\end{figure}

Applying existing DA methods to the EE task is more challenging than to translation or classification tasks, because we need to augment training data while keeping the event structure~(trigger and arguments) unchanged. Figure~\ref{fig:example} presents examples of three different DA methods applied to a sentence containing a ``\textit{Transport}'' event. The event is triggered by word ``\textit{left}'' and it has three arguments with different roles~(``\textit{Mike}'', ``\textit{this town}'' and ``\textit{yesterday}'').
As shown in Figure~\ref{fig:example}, it is infeasible to apply sentence-level DA methods such as BackTranslation~\citep{xie2020unsupervised}, because it may change the event structure~(change ``\textit{left}'' to ``\textit{went away from}''). 
Previous attempts on DA for such tasks typically use heuristic rules such as synonym replacement~\citep{dai2020analysis,Cai2020DataMT} or context-based words substitution with BERT~\citep{yang2019exploring}. Their idea is to replace adjunct tokens~(the tokens in sentences except triggers and arguments) with other tokens, and thus can ensure  the event structure is unchanged as much as possible. 
However, recent studies~\citep{ding2020daga,Yang2020GenerativeDA,Ye2022ZeroGenEZ}  find that such methods provide limited data diversity. 
In NLP, the diversity of data is mainly reflected in two aspects: expression diversity and semantic diversity~\citep{zhao2019data}. The Synonym Replacement and BackTranslation methods lack  semantic diversity, because they can only produce samples with similar semantics. The BERT-based method can only replace words and cannot change the syntax, so it cannot generate samples with a wide variety of expressions.
The lack of sufficient diversity may lead to greater overfitting or poor performance through training on examples that are not representative.

To this end, we present \textbf{Mask-then-Fill}, a flexible and effective data augmentation framework for event extraction. Our approach allows for more flexible manipulation of text and thus can generate more diverse data while keeping the original event structure unchanged as much as possible. Specifically, we first define two types of text fragments in a sentence: event-related fragments~(trigger and arguments) and adjunct fragments~(e.g. ``\textit{The police said}''). Then, we model DA for the EE task as a \textbf{Mask-then-Fill} process: we first randomly masks out an adjunct sentence fragment and then infills a variable-length text span with a fine-tuned infilling model~(T5)~\citep{raffel2020exploring}. The main advantage lies in that it can replace a fragment of arbitrary length in the text with another fragment of variable length, compared to the existing methods which can only replace a single word or a fixed-length fragment.

To the best of our knowledge, we are the first to augment training data for event extraction via text infilling.
We empirically show that the \textbf{Mask-then-Fill} framework improves performance for both classification-based~(EEQA) and generation-based~(Text2Event) event extraction models on a well-known EE benchmark dataset~(ACE2005). Especially, it demonstrates strong results in the low resource setting. We further investigate reasons for its effectiveness by introducing two metrics,  \textit{Affinity} and  \textit{Diversity}, and find that the data augmented by our approach have better diversity with less distribution shifts, achieving a good balance between diversity and distributional similarity.

\begin{figure} 
  \centering
  \includegraphics[width=0.8\linewidth]{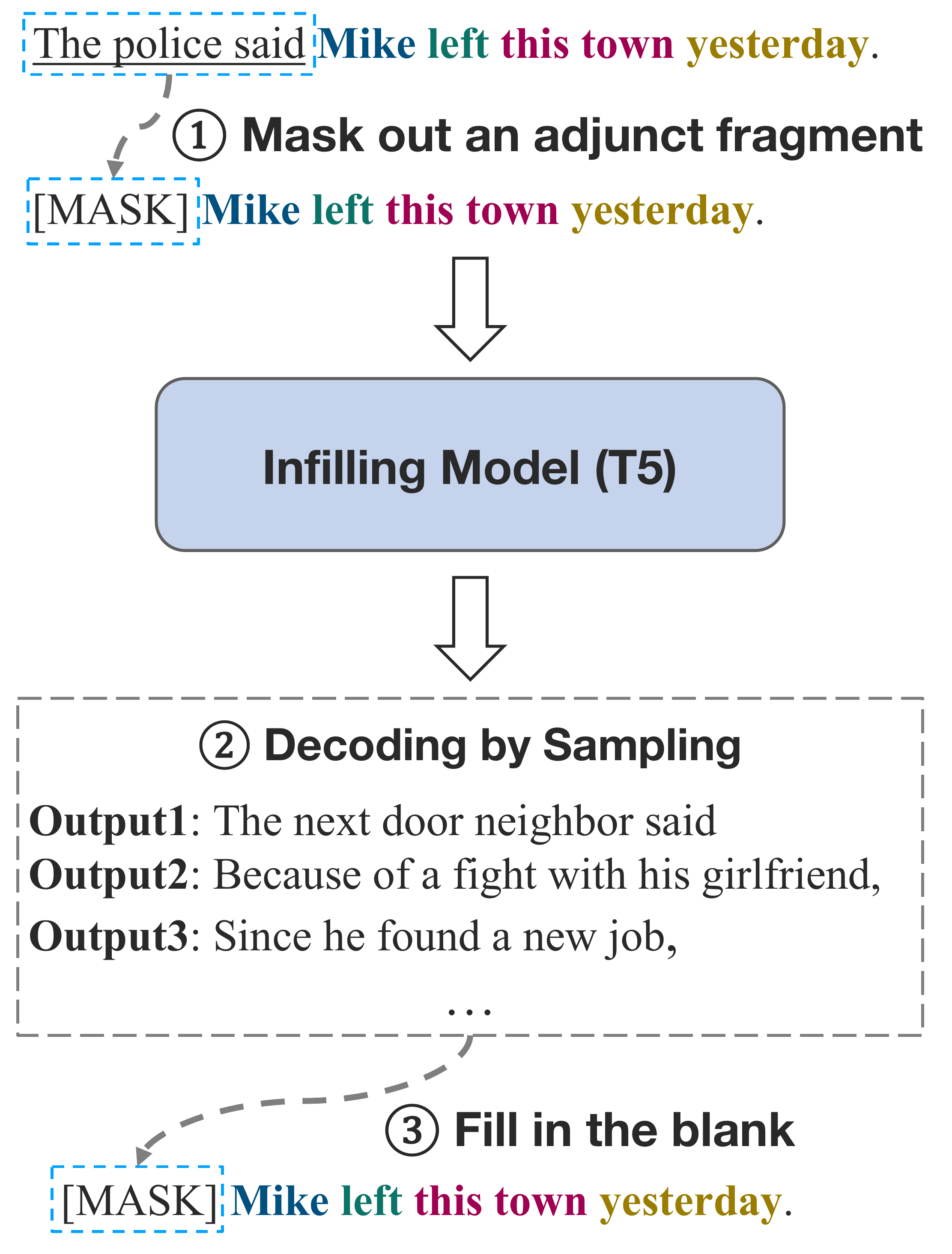}
  \caption{Overview of the proposed Mask-then-Fill framework.}
  \vspace{-1.2em}
  \label{fig:arch}
\end{figure}

\section{The Mask-then-Fill Framework}

Figure~~\ref{fig:arch} presents an overview of Mask-then-Fill framework.
The input sentence contains two types of text fragments: event-related fragments~(words with colors) and adjunct fragments~(underlined). Our idea is to rewrite the whole adjunct fragment instead of replacing some words, and the rewritten sentence fragment should fit the context and should not introduce new events.
To this end, we model DA for EE as a \textbf{Mask-then-Fill} process: we first randomly mask out an adjunct sentence fragment and then infills a variable-length text span with a fine-tuned infilling model. In the following, we describe in detail the Mask-then-Fill framework.

\paragraph{Mask out an adjunct fragment.} 
Given a prototype sentence $\bm{X}=\{x_1,\cdots,x_L\}$ of length $L$ from the training set, we first define an adjunct fragment as a set of non-overlapping spans of $x$ that do not contain the event triggers and arguments. We then replace one of the adjunct fragments with a [MASK] symbol. The incomplete sentence $\hat{\bm{x}}$  is a version of $\bm{x}$ with a fragment replaced with a [MASK] symbol.

\begin{table*}[htbp]
  \centering
  \small
      \begin{tabular}{|c|l|cccc|cccc|}
      \hline
      \multirow{2}{*}{\textbf{EE Model}}&\multirow{2}{*}{\textbf{DA Method}}&\multicolumn{4}{c|}{\textbf{Trig-ID+C~(F1)}}&\multicolumn{4}{c|}{\textbf{Arg-ID+C~(F1)}}\\
      \cline{3-10}
      &&S&M&L&F&S&M&L&F\\
      \hline
      \multirow{5}{*}{Text2Event}
        & No Augmentation & 45.44 & 59.75 & 63.55 & 67.06 & 22.05 & 36.04 & 40.35 & 49.30 \\ 
        & Synonym Replacement & \underline{49.14} & \underline{61.96} & 63.73 & \textbf{69.09} & \underline{27.71} & \underline{39.64} & \underline{43.63} & \underline{48.95} \\ 
        & BERT & 48.66 & 60.75 & 63.81 & 68.33 & 26.71 & 38.75 & 41.44 & 48.41 \\ 
        & Span-BackTranslation & 47.91 & 61.54 & \underline{64.59} & 67.58 & 26.68 & 38.18 & 43.39 & 47.93 \\ 
        & Ours (t5-small) & \textbf{52.32} & \textbf{63.38} & \textbf{67.25} & \underline{69.03} & \textbf{28.68} & \textbf{39.79} & \textbf{44.73} & \textbf{50.29} \\ 
       \hline
      
      \multirow{5}{*}{EEQA}
        & No Augmentation & 48.05 & 64.20 & 64.06 & 67.13 & 39.81 & 56.30 & 59.27 & 61.93 \\ 
        & Synonym Replacement & \textbf{54.86} & \underline{64.03} & 65.71 & 68.05 & \underline{42.99} & 56.62 & 56.40 & 61.50 \\ 
        & BERT & 53.61 & 63.23 & \underline{65.90} & 68.35 & 38.80 & 52.82 & \textbf{59.49} & \textbf{61.62} \\ 
        & Span-BackTranslation & 53.26 & 62.64 & 65.46 & \underline{68.40} & 42.47 & \underline{56.64} & 55.87 & \underline{61.55} \\ 
        & Ours (t5-small) & \underline{54.80} & \textbf{64.37} & \textbf{67.33} & \textbf{69.62} & \textbf{47.67} & \textbf{56.70} & \underline{58.52} & \textbf{61.62} \\ 

      \hline
      \end{tabular}
      \caption{Results on trigger extraction and argument extraction using different subsets of the training data. The best results  are marked in bold, and the second best is underlined.}
      \vspace{-1.5em}
      \label{tab:main_results}
\end{table*}

\paragraph{Blank Infilling Model.}
We formulate our blank infilling process as the task of predicting the missing span of text which is consistent with the preceding and subsequent text. Figure~\ref{fig:arch} gives an example with an incomplete input sentence $\widetilde{\bm{x}}$, where the [MASK] is a placeholder for a blank, which has masked out multiple tokens. Our goal is to predict only the missing span $\bm{y}$ which will replace the [MASK] token in $\widetilde{\bm{x}}$.
Therefore, the infilling task can be cast as learning $p(\bm{y}|\widetilde{\bm{x}})$.

To train our infilling model, we fine-tune a pretrained sequence-to-sequence model T5~\citep{raffel2020exploring} on the Gigaword corpus~\citep{graff2003english}, which is from similar domains as the event extraction dataset ACE2005 adopted by our work. Given a corpus consisting of plain sentences, we first produce large pools of infilling examples and then train the T5 model on these examples. For a given complete sentence $\bm{x}$ from the training corpus, we generate an infilling example $\widetilde{\bm{x}}$ with the following procedure: (1) randomly sample a span length $l$ from the range of $[1,\mathrm{min}(10,l)]$; (2) split the sentence into $l$ spans; (3) randomly select a span to be replaced with a [MASK] symbol. The replaced span is used as the target $\bm{y}$. We then fine-tune the T5 model on these infilling examples, yielding the model of the form $p_{\theta}(\bm{y}|\widetilde{\bm{x}})$.


\paragraph{Fill in the blank.}
Once trained, the infilling model can be used to take the incomplete sentence $\widetilde{\bm{x}}$, containing one missing span, and return a predicted span $\bm{y}$. We then replace the [MASK] token in $\widetilde{\bm{x}}$ with the predicted span  $\bm{y}$ to generate an augmented example. Note that we can produce large pools of augmented samples using top-k sampling.

\section{Experimental Setup}

\paragraph{Dataset.}
We empirically evaluate our proposed data augmentation method for event extraction on the ACE2005 corpus\footnote{https://catalog.ldc.upenn.edu/LDC2006T06}
with the same train-dev-test split and preprocessing step as previous works~\citep{zhang2019joint,wadden2019entity}.

We simulate a low-resource setting by randomly sampling 1,000, 4,000 and 8,000 examples from the training set to create the small, medium, and large training sets~(denoted as \textbf{S}, \textbf{M}, \textbf{L} in Table~\ref{tab:main_results}, whereas the complete training set is denoted as \textbf{F}). We only augment the training data and keep the dev set and test sets unchanged. 

\paragraph{Evaluation Metrics.}
Following the previous works~\citep{du2020event,lu2021text2event} on event extraction, we adopt the same evaluation criteria defined in \citet{li2013joint}: (i) An event trigger is correctly identified and classified~(\textbf{Trig-ID+C}) if its offsets match a gold trigger and its event type is also correct. (ii) An argument is correctly identified and classified~(\textbf{Arg-ID+C}) if its offsets and event type match a gold argument and its event role is also correct.

\paragraph{Event Extraction Models.}

In our study, we consider two representative models for event extraction:

\begin{itemize}[leftmargin=*,noitemsep,nolistsep]
  \item \textbf{Text2Event}~\citep{lu2021text2event} is a framework to solve the event extraction task by casting it as a SEQ2SEQ generation task.
  All triggers, arguments, and their labels are generated as natural language words.
  \item \textbf{EEQA}~\citep{du2020event} formulates the event extraction task as a question answering task. They develop two BERT-based QA models – one for event trigger detection and the other for argument extraction. 
\end{itemize}

\paragraph{Comparison Methods.}
We compare our proposed data augmentation method \textbf{Ours~(t5-small)} with three baselines: (1) \textbf{Synonym Replacement} replaces adjunct tokens with one of their synonyms retrieved from WordNet~\citep{miller1995wordnet} at random; (2) \textbf{BERT} replaces adjunct tokens with others randomly drawn according to the pretrained BERT's distribution; (3) \textbf{Span-BackTranslation}: Inspired by~\citet{yaseen2021data}, we only ``back translate'' randomly selected adjunct spans to prevent the model from changing the event structure.

\paragraph{Hyperparameters.}
For all data augmentation methods, we tune the number of augmentation samples per training sample from a list of numbers: $\{1,3,6,10\}$.

\section{Results and Analysis}

\paragraph{Main Results.}
The main results are presented in Table~\ref{tab:main_results}, where we use two EE models~(\textbf{Text2EVent} and \textbf{EEQA}) to test the performance of different DA methods in both low-resource~(\textbf{S}, \textbf{M} and \textbf{L}) and normal~(\textbf{F}) settings. As shown in the table, we observe that Ours~(t5-small) achieves the best overall performance among all DA methods on both trigger extraction~(F1) and argument extraction~(F1). 
Using our DA method gives the best results for the Text2event model on 7 out of 8 datasets. For the EEQA model, our method achieves the best results on 6 out of 8 datasets, where the difference between our method and the best results on \textbf{Trig-S} and \textbf{Arg-L} is very small, with only 0.06 and 0.97 points difference between them, respectively. Particularly, our methods demonstrates strong results in the low-resource setting. 
Using our DA  gives the Text2Event model a performance improvement of 15.14\% and 30.07\% on \textbf{Trig-S} and  \textbf{Arg-S}, respectively.
 
We also notice that as the amount of data increases, the improvement from all DA method decreases, and in some cases~(EEQA model on \textbf{Arg-L} and \textbf{Arg-F}), there is even a slight decrease in performance. 
In the case of more data, the model may overfit if the augmented data are just some similar samples rather than data with large variations.

\begin{table}[htbp]
  \centering
  \small
      \begin{tabular}{|l|ccc|}
      \hline
      \textbf{DA Method}&\textbf{Affinity}&\textbf{Dist-1}&\textbf{Dist-2}\\
      \hline
      Synonym Replacement & -0.118 & 0.400 & \underline{0.523} \\
        BERT &\textbf{-0.082} & 0.374 & 0.496 \\ 
        Span-BackTranslation & -0.155 & \underline{0.407} & 0.513 \\ 
        Ours~(t5-small) & \underline{-0.086} & \textbf{0.488} & \textbf{0.612} \\
      \hline
      \end{tabular}
      \caption{Results on \textit{Affinity} and \textit{Diversity}. The best results are marked in bold. The second best is underlined.}
      \vspace{-1.5em}
      \label{tab:affinity_diversity}
      
\end{table}
\paragraph{Affinity and Diversity.}
Inspired by \citet{gontijo2020affinity}, we further investigate reasons for its effectiveness by introducing two metrics,  \textit{Affinity} and \textit{Diversity}, where \textit{Affinity} quantifies how augmentation shifts data distribution and \textit{Diversity} measures the complexity of the augmented data. 
We measure \textit{Affinity} by computing the difference between the loss of a model trained on the original training set and tested on the original example, and the loss of the same model tested on an augmented example. We use the Dist-1/2 metric~\citep{celikyilmaz2020evaluation}, commonly used in text generation, to assess the \textit{Diversity} of the augmented data. For implementation details of two metrics, see Appendix.

We first construct a new test set by generating a new sample for each data in the test set. We then calculate the \textit{Affinity} and Dist-1/2 scores between the new data set and the original data set, respectively. As shown in Table~\ref{tab:affinity_diversity}, it is clear that the data augmented by our DA method have better diversity with less distribution shifts,
obtaining a balance between diversity and distributional similarity.


\begin{figure} 
  \centering
  \includegraphics[width=0.9\linewidth]{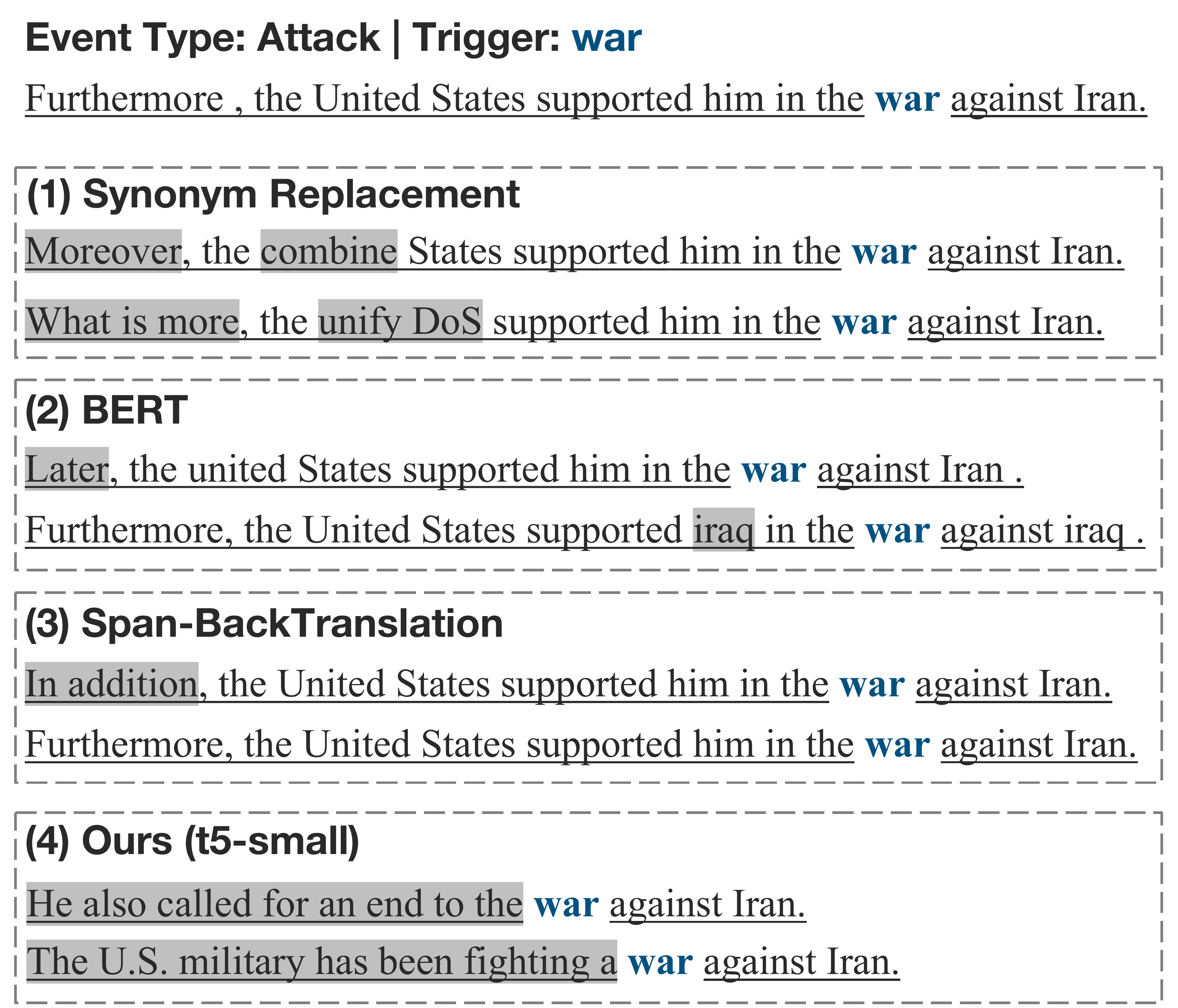}
  \caption{Augmented examples of four different DA methods. Given a sentence containing an ``Attack'' event triggered by the word "war", we generate two new samples for each DA method. The parts of the new sample that differ from the original are colored in gray.}
  \label{fig:case_study}
  \vspace{-1.5em}
\end{figure}

\paragraph{Case Study.}
Figure~\ref{fig:case_study} presents examples generated by different DA methods. Given a sentence containing an ``Attack'' event triggered by the word "war", we generated two new samples for each DA method, and the parts of the new sample that differ from the original are colored in gray. Obviously, The  synonym replacement based on WordNet cannot avoid introducing some words that do not fit the context~(e.g ``unify'' and ``DoS''), while the BERT-based word replacement can consider the context better. However, they both provide limited diversity. BackTranslation method performs even worse in terms of data diversity. Its generated data differs very little from the original sentence. Finally, 
compared with the original sentences, the new samples generated by our method are more fluent and more different in expression and semantics. Therefore, it not only generates data that fits the context better, but also provides better diversity.

\section{Conclusion}
In this paper, we present \textbf{Mask-then-Fill}, a flexible and effective data augmentation framework for event extraction. Our approach allows for more flexible manipulation of text and thus can generate more diverse data while keeping the original event structure unchanged.
The main advantage lies in that it can replace a fragment of arbitrary length in the text with another fragment of variable length. 
We empirically show that the \textbf{Mask-then-Fill} framework improves performance for both \textbf{EEQA} and \textbf{Text2Event} EE models on the ACE2005 dataset. It demonstrates particularly strong results in the low-resource setting. Our further analysis shows that it achieves a good balance between diversity and distributional similarity.

\section*{Limitations}
\label{sec:disscussion}
This paper presents a flexible and effective data augmentation framework for event extraction tasks. Here, we note some of \textbf{Mask-then-Fill} framework's limitations. First, performance gains can be marginal when data is sufficient. We believe this approach has much room for improvement in generating more diverse data. In this work, we select only one adjunct fragment at a time for modification, and modifying multiple adjunct fragments in an event mention can further enhance the diversity of the generated data. Second, currently this method can only replace one fragment at a time. This makes it easier to control the properties of the generated fragments, such as length or style. It is possible to modify multiple fragments at the same time using some existing techniques~\citep{donahue2020enabling,du2022glm,chen2022paraphrase}. This approach is more efficient, but it is prone to generate incoherent augmented samples and thus introduce more noise. A possible approach to solve this problem is to design some sample selection strategies.

\section*{Acknowledgements}
This work was partially supported by the National Natural Science Foundation of China 62006062 and 62176076, Shenzhen Foundational Research Funding JCYJ20200109113441941, JCYJ20210324115614039, The Major Key Project of PCL2021A06, Guangdong Provincial Key Laboratory of Novel Security Intelligence Technologies 2022B1212010005.

\bibliography{custom,reference}
\bibliographystyle{acl_natbib}

\appendix







\section{Affinity and Diversity}
Inspired by \citet{gontijo2020affinity} and \citet{arora2021types}, we proposed to use a calibration method to quantify how augmentation shifts data. They all note that a trained model is often sensitive to the distribution of the training data.

Given the original example $\bm{x}$ and one of its augmented example $\bm{x}^+$, we measure distribution shifts by computing the difference between the loss of a model trained on the original training set and tested on the original example, and the loss of the same model tested on an augmented example:
\begin{equation}
  \tau_{\alpha} = \ell(M,\bm{x})-\ell(M,\bm{x}^+),
\end{equation}
where $M$ is an EE model trained on the original training set and $\ell(M,\bm{x}^+)$ denotes the model’s validation loss when evaluated on the augmented example $\bm{y}$.

We use the Dist-1/2 metric~\citep{celikyilmaz2020evaluation}, commonly used in text generation, to assess the \textit{Diversity} of the augmented data.

\section{Implementation Details}

\begin{table}[h]
\setlength{\abovecaptionskip}{0.1cm}
\setlength{\belowcaptionskip}{-0.0cm}
\centering
\begin{tabular}{lc}
\toprule
\textbf{Parameter}                & \textbf{Value} \\ 
\midrule
Training Epochs                   & 3              \\
Optimizer                         & AdamW          \\
Batch Size                        & 64             \\
Learning rate             & 1e-5           \\
Seed                      & 1024            \\ 
Top-k                      & 100            \\ 
Top-p                      & 0.7            \\ 
Beam Size                      & 5            \\ 
\bottomrule
\end{tabular}
\caption{Implementation details of our infilling model~(t5-small).}
\end{table}

\begin{table}[h]
\setlength{\abovecaptionskip}{0.1cm}
\setlength{\belowcaptionskip}{-0.0cm}
\centering
\begin{tabular}{lc}
\toprule
\textbf{Parameter}                & \textbf{Value} \\ 
\midrule
Training Epochs                   & 30              \\
Optimizer                         & AdamW          \\
Batch Size                        & 64             \\
Learning rate             & 5e-5           \\
Seed                      & 1024            \\ 
\bottomrule
\end{tabular}
\caption{Implementation details of Text2Event.}
\end{table}

\begin{table}[h]
\setlength{\abovecaptionskip}{0.1cm}
\setlength{\belowcaptionskip}{-0.0cm}
\centering
\begin{tabular}{lc}
\toprule
\textbf{Parameter}                & \textbf{Value} \\ 
\midrule
Training Epochs                   & 30              \\
Optimizer                         & AdamW          \\
Batch Size                        & 64             \\
Learning rate             & 4e-5           \\
Seed                      & 1024            \\ 
nth query & 5 \\
\bottomrule
\end{tabular}
\caption{Implementation details of EEQA.}
\end{table}

\end{document}